\documentclass{article}
\usepackage{amsmath,graphicx,mlspconf}
\usepackage{xcolor}
\usepackage{hyperref}

\copyrightnotice{U.S.\ Government work not protected by U.S.\ copyright}

\copyrightnotice{979-8-3503-2411-2/25/\$31.00 {\copyright}2025 Crown}

\copyrightnotice{979-8-3503-2411-2/25/\$31.00 {\copyright}2025 European Union}

\copyrightnotice{979-8-3503-2411-2/25/\$31.00 {\copyright}2025 IEEE}

\toappear{2025 IEEE International Workshop on Machine Learning for Signal Processing, Aug.\ 31-- Sep.\ 3, 2025, Istanbul, Turkey}

\newcommand{\eq}[1]{(\ref{#1})}

\newcommand{\mybold}[1]{\boldsymbol{#1}}
\newcommand{\bi}{\begin{itemize}}
\newcommand{\ei}{\end{itemize}}

\title{CONVOLUTIONAL SPIKING-BASED GRU cell FOR SPATIO-TEMPORAL DATA}
\name{%
  Yesmine Abdennadher\hfill
  Eleonora Cicciarella\hfill
  Michele Rossi \thanks{This work has been supported by the EU H2020 MSCA ITN project
Greenedge (grant no. 953775).}
}
\address{%
  Department of Information Engineering, University of Padova\\
  \{yesmine.abdennadher, michele.rossi\}@unipd.it,\; eleonora.cicciarella@phd.unipd.it
}

\begin{document}

\maketitle

\begin{abstract}
Spike-based temporal messaging enables SNNs to efficiently process both purely temporal and spatio-temporal time-series or event-driven data. Combining SNNs with Gated Recurrent Units (GRUs), a variant of recurrent neural networks, gives rise to a robust framework for sequential data processing; however, traditional RNNs often lose local details when handling long sequences. Previous approaches, such as SpikGRU, fail to capture fine-grained local dependencies in event-based \mbox{spatio-temporal} data. In this paper, we introduce the Convolutional Spiking GRU (CS-GRU) cell, which leverages convolutional operations to preserve local structure and dependencies while integrating the temporal precision of spiking neurons with the efficient gating mechanisms of GRUs. This versatile architecture excels on both temporal datasets (NTIDIGITS, SHD) and spatio-temporal benchmarks (MNIST, DVSGesture, CIFAR10DVS). Our experiments show that CS-GRU outperforms state-of-the-art GRU variants by an average of 4.35\%, achieving over 90\% accuracy on sequential tasks and up to 99.31\% on MNIST. It is worth noting that our solution achieves 69\% higher efficiency compared to SpikGRU. The code is available at: \href{https://github.com/YesmineAbdennadher/CS-GRU}{https://github.com/YesmineAbdennadher/CS-GRU.}
\end{abstract}

\section{INTRODUCTION}
\label{sec:intro}

Spatiotemporal data are at the heart of many modern applications, ranging from audio and video analysis to gesture and event recognition. The time-changing nature of such data types requires models that not only capture long-range dependencies but also preserve localized structure. Spiking Neural Networks (SNNs) with their biologically inspired spike-based communication patterns, yield unique advantages in energy efficiency and temporal precision. Moreover, Gated Recurrent Units (GRUs) are particularly good at capturing sequence dynamics, though sometimes at the cost of missing to capture important local details. Although GRU cells have been extremely successful, typical recurrent neural networks and spiking neuron models like Leaky-Integrate and Fire (LIF) neurons are intrinsically limited in managing complex sequential inputs. In long sequences, critical local information gets diluted, leading to suboptimal performance on both temporal and spatiotemporal classification tasks. This limitation is particularly relevant when spatial relations are central to making good predictions, i.e., for a model that not only depends on long-term dynamics but that also retains local structure information. In this papepr we propose the Convolutional Spiking GRU (CS-GRU) cell to meet these needs. CS-GRU unites convolutional processing with spiking neural dynamics and GRU gating, enabling the model to preserve local dependencies while robustly extracting long-term temporal patterns. Inspired by the event-driven, efficient communication in biological neural networks, the CS-GRU not only excels on temporal data but also on spatiotemporal tasks, where local structural features are of primary importance. Apart from its immediate performance benefit, our CS-GRU block also has the potential to serve as a fundamental building block in larger deep learning systems. Its design allows for the replacement of conventional neuron models, such as LIF and Cuba LIF, in intricate architectures, including ensembles, ResNets, and VGG networks. This flexibility can lead to more efficient, brain-inspired models that bridge the gap between computational efficiency and biological plausibility, offering new research opportunities as well as more effective models for practical applications. 

The novel contributions of our work are as follows:
\bi
\item Starting from the state-of-the-art SpikGRU design~\cite{dampfhoffer2022investigating}, we introduce a few key modifications to their GRU neurons, by changing the corresponding gating mechanism, and using the convolution operation.
\item Additional enhancements are attained by replacing the Heaviside activation function of spiking networks with the arctangent, and using surrogate gradients.
\item Selected spiking architectures are thoroughly tested by assessing their performance with static and dynamic (temporal data) datasets, proving the superiority of our final CS-GRU design. In particular, the proposed CS-GRU obtains average improvements of $4.35$\% against SpikGRU, by achieving accuracies higher than $90$\% for challenging sequential datasets, and significantly higher accuracies for spatio-temporal data, e.g., up to $99.31$\% for the MNIST dataset.
\ei

The rest of the paper is organized as follows. In Section~\ref{sec:related_work}, we review the related work on spiking neural networks, with a focus on the latest GRU‑based designs. In Section~\ref{sec:Preliminaries}, we present the necessary preliminaries and notation. In section~\ref{sec:method}, we present our convolutional spiking GRU (CS-GRU) architecture, by discussing its design and all the modifications that we introduced. Experimental results, for temporal and spatio-temporal data, are reported in Section~\ref{sec:exp}. Finally, conclusions and directions for future research are outlined in Section~\ref{sec:conclusion}.

\section{RELATED WORK}
\label{sec:related_work}
Spiking Neural Networks (SNNs), the third generation of neural networks, generalize earlier models by simulating biological neurons that exchange information using discrete, time-encoding spikes~\cite{maass1997networks,gerstner2002spiking}. Unlike traditional continuous-activation networks, SNNs are event-driven, allowing for native support of temporal and spatiotemporal patterns. The event-driven nature constrains unnecessary computation and energy usage, rendering SNNs appropriate for low-resource settings like edge and wearable devices. Their effectiveness with temporal and spatiotemporal data, along with advances in neuromorphic hardware, enable real-time, power-efficient computation. Various spiking neuron models have been presented to capture different aspects of biological neural activity. The Hodgkin–Huxley (HH) model~\cite{hausser2000hodgkin} is a biophysical model with multiple ion channels to capture the complex dynamics of neuronal spiking.
The Leaky Integrate-and-Fire (LIF) model~\cite{brunel2007lapicque}, often idealized as a leaky capacitor charging and discharging as a function of time, is a simpler approximation of neuronal dynamics. Building on the LIF model, the Adaptive Exponential Integrate-and-Fire (AdEx) model~\cite{brette2005adaptive} adds an exponential term to capture adaptation of firing rates as a function of time. Additionally, the Spike Response Model (SRM) characterizes the post-spike response properties of neurons, including the refractory period and the shape of the post-spike potential. Together, these models comprise a set of approaches that balance biological accuracy and computational convenience for modeling neural activity. Recent work has pushed further along the lines of combining conventional recurrent neural network topologies with models of neurons so that they capture biological spiking more realistically. For instance, the Spiking Recurrent Cell~\cite{de2024spike} modifies the temporal dynamics of an ordinary GRU so that they become appropriate for event-based computing. Recently, a variant of the LIF model, referred to as SpikGRU~\cite{dampfhoffer2022investigating}, has been proposed, which combines spike-based processing with GRU cell design. These techniques bring together the time-dependent, discrete spiking dynamics of biological neurons and adaptive gate mechanisms of GRUs.

\section{PRELIMINARIES}
\label{sec:Preliminaries}

\subsection{Current-based leaky integrate and fire (Cuba-LIF)}
\label{ssec:cubaLIF}

 The Leaky Integrate-and-Fire (LIF)~\cite{brunel2007lapicque} model is the most widely adopted for spiking neural networks (SNNs). LIF mimics the behavior of biological neurons by reproducing the gradual summation of input potentials up to a threshold, at which a spike is generated at its ouput and the accumulated (local) potential is reset. Worth of note, is the so called Cuba-LIF model~\cite{gerstner2014neuronal}, a LIF variant that employs current-based synapses. In Cuba-LIF, an input current \(i\) (carrying input spikes) is integrated before modifying the membrane potential \(v\). In this paper we denote by $i_t^\ell$ and $v_t^\ell$ the input current and the local membrane potential at network layer $\ell$ and time $t$, respectively. \(\odot\) is the elementwise product, and we use bold symbols for vectors and matrices.

As shown in \eq{eq:spike}, the integration of the input current \(\mybold{i}^\ell\) is regulated by a decaying parameter \(\alpha\), by matrices \(\mybold{W}^\ell\), \(\mybold{U}^\ell\), and vector \(\mybold{b}^\ell\). The membrane potential \(\mybold{v}_t^\ell\) is defined as a linear combination of the previous neuron state \(\mybold{v}_{t-1}^\ell\), moderated by a constant \(\beta\), and the current input \(\mybold{i}_t^\ell\), weighted by constant \(1-\beta\). $\mybold{s}^\ell_t$ represents the neuron output at time $t$, which is calculated using the Heaviside step function $H(\cdot)$ (applied elementwise), i.e., the neuron fires when the local potential \(\mybold{v}_t^\ell\) reaches a threshold \(\mybold{v}_{\mathrm{th}}\).
\begin{equation}
\begin{aligned}
\mybold{i}_t^\ell &= \alpha \odot \mybold{i}_{t-1}^\ell 
       + \mybold{W}^\ell\,\mybold{s}_{t}^{\ell-1} 
       + \mybold{U}^\ell\,\mybold{s}_{t-1}^{\ell} 
       + \mybold{b}^\ell, \\
\mybold{v}_t^\ell &= \beta \odot \mybold{v}_{t-1}^\ell 
       + (1 - \beta) \odot \mybold{i}_t^\ell 
       - \mybold{v}_{\mathrm{th}}\,\mybold{s}_{t-1}^{\ell}, \\
\mybold{s}_t^\ell &= H\bigl(\mybold{v}_t^\ell - \mybold{v}_{\mathrm{th}}\bigr).
\end{aligned}
\label{eq:spike}
\end{equation}
Note that the term \( \mybold{v}_{\mathrm{th}}\,\mybold{s}_t^{\ell-1} \), which is subtracted from the second equation corresponds to the reset mechanism for the membrane potential $\mybold{v}_t^\ell$, whenever the membrane threshold $\mybold{v}_{\mathrm{th}}$ is reached in the previous time step $t-1$ (mimicking the membrane discharge after firing a spike). 

\subsection{Gated Recurrent Units (GRUs)}

Recurrent neural networks (RNNs) architectures can predict time-series or sequential data and are therefore particularly suitable for use where the input is of different lengths, such as with biological signals, natural language, and video~\cite{medsker2001recurrent}. RNNs employ hidden states that allow previous outputs to be utilized as inputs, and the network can learn temporal relationships within the data. However, regular RNNs have a tendency to be affected by the {\it vanishing gradient} problem, which prevents them from learning long-term dependencies; to counteract this limitation, modifications like Long Short-Term Memory (LSTM) networks and Gated Recurrent Units (GRUs)~\cite{dey2017gate} were presented. GRUs offer a leaner alternative by combining the forget and input gates into an update gate, hence reducing the number of parameters with respect to LSTMs without compromising performance. The GRU equations are reported below in \eq{eq:GRU}, where $\mybold{x}_t^\ell$ is the input at layer $\ell$ and time $t$ and $\mybold{h}_t^\ell$ is the GRU cell state, which also corresponds to the cell output at time $t$.
\begin{equation}
\begin{aligned}
\mybold{z}_t^\ell &= \sigma\bigl(\mybold{W}_z^\ell \mybold{x}_t^\ell + \mybold{U}_z^\ell \mybold{h}_{t-1}^\ell + \mybold{b}_z^\ell\bigr), \\
\mybold{r}_t^\ell &= \sigma\bigl(\mybold{W}_r^\ell \mybold{x}_t^\ell + \mybold{U}_r^\ell \mybold{h}_{t-1}^\ell + \mybold{b}_r^\ell\bigr), \\
\tilde{\mybold{h}}_t^\ell &= \tanh\bigl(\mybold{W}_h^\ell \mybold{x}_t^\ell + \mybold{U}_h^\ell \bigl(\mybold{h}_{t-1}^\ell \odot \mybold{r}_t^\ell\bigr) + \mybold{b}_h^\ell\bigr), \\
\mybold{h}_t^\ell & = \mybold{z}_t^\ell \odot \mybold{h}_{t-1}^\ell + \bigl(1 - \mybold{z}_t^\ell\bigr) \odot \tilde{\mybold{h}}_t^\ell,
\end{aligned}
\label{eq:GRU}
\end{equation}
where \(\sigma(\cdot)\) represents the sigmoid function, symbols \(\mybold{W}\) and \(\mybold{b}\) refer to  weight matrices and bias vectors, \( \tanh(\cdot) \) denotes the hyperbolic tangent function.

In a GRU, the reset gate \(\mybold{r}_t\) determines how much of the previous state \(\mybold{h}_{t-1}\) to consider when updating with new input. This is achieved by linearly interpolating the current input \(\mybold{x}_t\) and the previous state \(\mybold{h}_{t-1}\) using the respective weights and biases, and then squashing the result through an activation function (such as \( \tanh(\cdot) \)) to produce a value between $0$ and $1$. The update gate \(\mybold{z}_t\), meanwhile, controls how much the state is updated at the current time step $t$, by balancing how much information from the past is carried forward versus how much new inputs are incorporated into it. Following that, the candidate  state equation \( \tilde{\mybold{h}}_t \) is introduced. The candidate state is computed by applying a non-linear activation, most commonly the hyperbolic tangent, to a linear combination of the current input \(\mybold{x}_t\) and the previous state \(\mybold{h}_{t-1}\), weighted by the reset gate \( \mybold{r}_t \). Finally, considering the update gate \( \mybold{z}_t \), the final state \( \mybold{h}_t \) is obtained by combining the previous state and the candidate state. In the following, the GRU is used as the key building block for the proposed CS-GRU model.

\label{ssec:gru}
\subsection{Spiking-based GRU (SpikGRU)}
\label{ssec:spikgru}

SpikGRU~\cite{dampfhoffer2022investigating} extends the Cuba-LIF model by incorporating GRU-style gating mechanisms within a spiking neural network (SNN) framework. Designed specifically for audio tasks, where accurately modeling temporal dynamics is essential, SpikGRU enhances Cuba-LIF by combining traditional spiking properties with the adaptive features of GRUs. In this model, an update gate \(\mybold{z}\) is introduced; instead of relying on a fixed parameter (such as \(\beta\)), the gate is implemented through a sigmoid function, The gate output is determined using the cell input at time $t$ and output at time $t-1$, with a dedicated set of parameters, \(\mybold{W}_z\), \(\mybold{U}_z\), and \(\mybold{b}_z\), as follows:
\begin{equation}
\begin{aligned}
\mybold{i}_t^\ell &= \alpha \odot \mybold{i}_{t-1}^\ell + \mybold{W}_i^\ell\, \mybold{s}_{t}^{\ell-1} + \mybold{U}_i^\ell\,\mybold{s}_{t-1}^\ell + \mybold{b}_i^\ell, \\
\mybold{z}_t^\ell &= \sigma\bigl(\mybold{W}_z^\ell\,\mybold{s}_t^{\ell-1} + \mybold{U}_z^\ell\,\mybold{s}_{t-1}^\ell + \mybold{b}_z^\ell\bigr), \\
\mybold{v}_t^\ell &= \mybold{z}_t^\ell \odot \mybold{v}_{t-1}^\ell + \bigl(1 - \mybold{z}_t^\ell\bigr)\odot \mybold{i}_t^\ell \;-\; \mybold{v}_{\mathrm{th}}\,\mybold{s}_{t-1}^\ell, \\
\mybold{s}_t^\ell &= H\bigl(\mybold{v}_t^\ell - \mybold{v}_{\mathrm{th}}\bigr).
\end{aligned}
\label{eq:neural_mech}
\end{equation}

\section{Convolutional Spiking GRU (CS-GRU)}
\label{sec:method}
In this paper, we present CS-GRU. In this new architectural design the local structure of sequential data is preserved thanks to the convolution operation. The convolution picks up on local patterns, e.g., short durations of audio events in spectrograms or transient visual cues, and this localized information is then integrated over time by the recurrent architecture. In this way, the model gains the ability not only to detect subtle invariances or changes in the input, but also weight sharing is attained across the spatial (thanks to convolution) and temporal (thanks to the GRU design) dimensions, to attain better generalization on challenging sequential tasks. Spurred by these benefits, and inspired by the temporal precision and energy efficiency of spiking neural dynamics, we built our convolutional spiking GRU, which integrates the localized feature extraction of convolutional layers with the robust temporal modeling of spiking GRUs. The design of our CS-GRU is explained in the following Sections~\ref{sec:ref}--\ref{sec:CS_GRU}.


\subsection{Adding a current gate (mod1)}
\label{sec:ref}

SpikGRU~\cite{dampfhoffer2022investigating} calculates the current \(\mybold{i}_t^\ell\) using a {\it constant} decay parameter \( \alpha \), see \eq{eq:neural_mech}, to retain information from the past. Here, to gain additional flexibility, a gate \( \mybold{r}_t \) is used in place of the constant \( \alpha \) from \eq{eq:neural_mech} (see first line). This gate is derived from the input spikes and provides a dynamic, time-dependent management on the retention of past information. As opposed to the fixed decay factor \( \alpha \), which applies the same degree of decay to all inputs regardless of their spiking patterns, the gate \( \mybold{r}_t \) can modulate its response based on whether its input activity is a spike burst or is extended over time. Hence, our first modification (referred to as ``mod1'') to the SpikGRU model amounts to replacing the first line of \eq{eq:neural_mech} with:
\begin{equation}
\begin{aligned}
\mybold{r}_t^\ell &= \sigma\bigl(\mybold{W}_r^\ell\, \mybold{s}_t^{\ell-1} + \mybold{U}_r^\ell\, \mybold{s}_{t-1}^\ell + \mybold{b}_r^\ell\bigr), \\
\mybold{i}_t^\ell &= \mybold{r}_t^\ell \odot \mybold{i}_{t-1}^\ell + \mybold{W}_i^\ell\, \mybold{s}_{t-1}^\ell + \mybold{U}_i^\ell\, \mybold{s}_t^{\ell-1} + \mybold{b}_i^\ell.
\end{aligned}
\end{equation}

\subsection{Membrane potential dependency (mod2)}
\label{sec:mem_pot}

In Cuba‐LIF, the membrane potential is directly computed based on current input and output of the preceding layer $\ell-1$, and thus current-based information as well as history information are encoded into the neuron state. SpikGRU avoids the use of the current input in the gating mechanism, however. Its update gate does not contain the current input as a direct reference any longer, and this alters the mechanism of incorporating new information, as compared to Cuba‐LIF. This change involves that while Cuba‐LIF combines the current and past indicators in an explicit way, SpikGRU's gating originates from other variables, which can influence its sensitivity to rapid changes in the input. For instance, SpikGRU gate \( \mybold{z}_t^\ell \) is defined as:
\begin{align}
\mybold{z}_t^\ell &= \sigma\bigl( \mybold{W}_z^\ell\,\mybold{s}_t^{\ell-1} + \mybold{U}_z^\ell \,\mybold{s}_{t-1}^\ell + \mybold{b}_z^\ell \bigr) .
\end{align}
In this equation, however, only the membrane potential is accounted for. Instead, we propose (second modification to SpikGRU, ``mod2'') to use current $\mybold{i}_t$ for the gate computation, as it follows:
\begin{equation}
\begin{aligned}
\mybold{z}_t^\ell &= \sigma\bigl(\mybold{W}_z^\ell\, \mybold{i}_t^\ell + \mybold{U}_z^\ell \, \mybold{s}_{t-1}^\ell + \mybold{b}_z^\ell \bigr) , 
\end{aligned}
\end{equation}
Through this modification, the gate \( \mybold{z}_t \) becomes context-sensitive, as it allows the neuron model to contextually modulate the effect of past information based on the current state at time $t$, permitting a more sensitive and time-aware control. We empirically found this modification to make a remarkable difference when dealing with rapidly changing or bursty spiking activity, where the current variable $\mybold{i}_t$ can offer valuable clues for selective recall or forgetting of past inputs.

\subsection{Using convolutions (mod3)}
\label{sec:convolution}

Our last modification to SpikGRU (``mod3'') is to use the convolution operation to extend the SpikGRU unit, by replacing the Fully-Connected (FC) linear product operations with convolutional ones. In SpikGRU, operations such as $\mybold{W} \mybold{s}_t$ are matrix multiplications that treat the input as a flat vector and largely disregard its spatial correlation structure. On the other hand, in a Convolutional GRU, these FC operations are replaced with convolutional filters, given as $\mybold{W} * \mybold{s}_t$,
where the convolution operator $*$ allows the model to learn local patterns along the spatial dimension that may go undetected by SpikGRU. 



\subsection{Arctan surrogate gradient (mod4)}
\label{sec:arctan}

The arctan surrogate offers a polynomially decreasing gradient that is always nonzero for all membrane potentials \cite{neftci2019surrogate}, eliminating "dead" areas because of triangular surrogate's precipitous, zero‑elsewhere slope, as used in SpikGRU \cite{dampfhoffer2022investigating}. And, being infinitely differentiable, delivers smoother loss terrain and more convergent updates, which in the real world is much more accurate on long-sequence and neuromorphic datasets than the error-prone triangular approximation.

\subsection{Convolutional Spiking GRU (CS-GRU)}
\label{sec:CS_GRU}

\begin{figure}[tb]
\centering
  \includegraphics[width=\columnwidth]{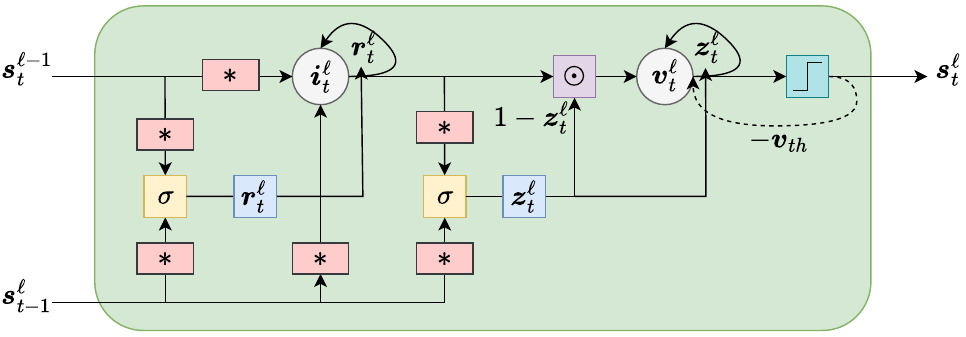}
  \caption{CS-GRU cell.}
  \label{fig:pipeline}
\end{figure}

The proposed CS-GRU model is obtained by concurrently using mod1, mod2, mod3 and mod4, as follows (see also Fig.~\ref{fig:pipeline}):
\begin{equation}
\begin{aligned}
\mybold{r}_t^\ell &= \sigma \bigl(\mybold{W}_r^\ell\, * \mybold{s}_t^{\ell-1} + \mybold{U}_r^\ell\, * \mybold{s}_{t-1}^\ell + \mybold{b}_r^\ell \bigr), \\
\mybold{i}_t^\ell &= \mybold{r}_t^\ell \odot \mybold{i}_{t-1}^\ell + \mybold{W}_i^\ell\, * \mybold{s}_{t}^{\ell-1} + \mybold{U}_i^\ell\, * \mybold{s}_{t-1}^\ell + \mybold{b}_i^\ell,\\
\mybold{z}_t^\ell &= \sigma\bigl(\mybold{W}_z^\ell\, * \mybold{i}_t^\ell + \mybold{U}_z^\ell\, * \mybold{s}_{t-1}^{\ell} + \mybold{b}_z^\ell\bigr), \\
\mybold{v}_t^\ell &= \mybold{z}_t^\ell \odot \mybold{v}_{t-1}^\ell + \bigl(1 - \mybold{z}_t^\ell\bigr) \odot \mybold{i}_t^\ell \;-\; \mybold{v}_{\mathrm{th}}\, \mybold{s}_{t-1}^\ell, \\
\mybold{s}_t^\ell &= H\bigl(\mybold{v}_t^\ell - \mybold{v}_{\mathrm{th}}\bigr).
\end{aligned}
\label{eq:neural}
\end{equation}

\section{Experiments}
\label{sec:exp}

\subsection{Temporal data}

For the purpose of classifying temporal data, we considered two speech recognition datasets. N-TIDIGITS is a spike dataset derived from audio recordings from the Dynamic Audio Sensor, a spiking silicon cochlea sensor, of the TIDIGITS audio set. The database includes $64$-channel recordings and comprises $11$ classes of the English digits ``one'' to ``nine'', plus ``oh'' and ``zero''. These digits were read by $111$ speakers for the training set and by $109$ speakers for the test set. The one-digit N-TIDIGITS has $2,464$ training samples and $2,486$ test samples. Additionally, every sample is trimmed to $1.25$ seconds, hence the sequence length is $250$ timesteps~\cite{anumula2018feature}. SHD~\cite{cramer2020heidelberg}, on the other hand, is a spiking version of the Heidelberg Digits sound dataset, produced through an audio-to-spiking mapping process that utilizes neurophysiologically inspired principles. The dataset maps $700$ channels to represent $20$ English and German spoken digit classes for $12$ speakers. It has $8,156$ training samples and $2,264$ test samples. The test set also includes recordings of two unseen speakers during training, along with an additional $5$\% of samples from other speakers. The resulting spike events are passed directly into the models over $250$ timesteps.

For our training procedure, we employ a single recurrent layer consisting of $128$ units, followed by a readout layer that integrates inputs via self-recurrence, as follows:
\begin{equation}
    \textbf{out}_t = \alpha\, \textbf{out}_{t-1} + \mybold{W}_{\rm out} \mybold{x}_t + \mybold{b}.
\end{equation}
We use a max-over-time loss similar to that employed by SpikGRU~\cite{cramer2020heidelberg}, which computes the maximum neuron activation in the readout layer over the time axis. We used backpropagation through time and the Adam optimizer~\cite{kingma2015adam} for training our model with a $0.001$ learning rate, $100$ epochs, and a batch size of $128$. Convolution is inherently designed to operate on multi-dimensional spatial data arrays. Therefore, we reshape our inputs to be in a \( C \times H \times W \) format. For example, the DVS Gesture dataset is reshaped to \( 1 \times 8 \times 8 \), and the SHD dataset is reshaped to \( 7 \times 10 \times 10 \).
We also downsample the DVS Gesture input by applying a \(2 \times 2\) max-pooling layer, and downsample the SHD input by using a \(3 \times 63\) convolution to reduce the spatial dimensions.
For the spiking behavior, we set the spiking threshold to $v_{\rm th}=1$ and use the scaled hyperbolic tangent (see ``mod4'') as the surrogate gradient~\cite{neftci2019surrogate} to mitigate the non-differentiability of the Heaviside activation.

Numerical results are reported in the following Table~\ref{tab:1x128_results}. On both SHD and N‑TIDIGITS, the vanilla (non-spiking) GRU baseline establishes high baselines (86.2\% and 84.5\%, respectively), and a minimal Cuba‑LIF spiking layer trails behind. SpikGRU improves over GRU on N‑TIDIGITS (to $86.2$\%) and on SHD (to $87.9$\%) datasets. Individual modifications ("mod1"–"mod4") are subsequently applied: some of them (mod2) achieve good results ($87.6$\% on SHD), while others alone (mod3) drastically reduce performance. Notably, however, using all four modifications leads to the highest accuracy across the entire collection of data, with $90.66$\% on N‑TIDIGITS and $91.96$\% on SHD.

\begin{table}[htp]
\centering
\caption{Accuracy (\%) of different $1\times 128$ networks on N-TIDIGITS, SHD.}
\label{tab:1x128_results}
\begin{tabular}{lcc}
\hline
\textbf{Model} & \textbf{NTIDIGITS} & \textbf{SHD}  \\
\hline
GRU       & $86.19 $ & $84.49$  \\

Cuba-LIF  & $81.25$ & $ 73.01 $           \\
SpikGRU   & $86.23$ & $87.89$  \\
SpikGRU-mod1   & $82.61$ & $84.71$  \\
SpikGRU-mod2   & $86.99$ & $87.63$  \\
SpikGRU-mod3   & $82.12$ & $90.06$  \\
SpikGRU-mod4   & $86.28$ & $85.77$  \\
SpikGRU-mod1-3   & $37.61$ &  $91.43$ \\
SpikGRU-mod2-3   & $69.55$ &  $40.50$ \\
SpikGRU-mod3-4   & $55.79$ &  $8.43$ \\
SpikGRU-mod1-2-3   & $ 73.9$ &  $89.48$ \\
SpikGRU-mod1-3-4  & $24.60$ &  $86.79$ \\
SpikGRU-mod2-3-4  & $17.30$ &  $86.7$ \\
SpikGRU-mod1-2-3-4   &\textbf{90.66}       & \textbf{91.96} \\
\hline
\end{tabular}
\end{table}

\subsection{Spatio-temporal data}

The MNIST dataset, traditionally composed of grayscale handwritten digit images, has been revised as a static dataset with an artificial temporal dimension. In our experiment, each image is elongated into a sequence that persists over $10$ timesteps using rate coding~\cite{kim2022rate}.
DVS128Gesture \cite{amir2017low} is an event-based dataset of a dynamic vision sensor (DVS) observing various hand and arm gestures. In our setup, we break the gesture streams into 10 timesteps, which provides us with a short but informative time window that maintains the high-speed dynamics of gestures. The CIFAR10 DVS dataset \cite{li2017cifar10} is derived from the original CIFAR10 images by converting them into an event-based representation through simulation of a neuromorphic camera or by employing a conversion algorithm. This mapping converts static images into sequences that mimic the event-driven sensor dynamics. In our method, each sample is organized into 10 timesteps. We adopt hyperparameters similar to those used for sequential data processing, but train for 300 epochs.
\begin{table}[h]
\centering
\caption{SpikGRU and CS-GRU accuracy on different datasets.}
\label{tab:comparison}
\begin{tabular}{lccc}
\hline
\textbf{} &  \textbf{MNIST } & \textbf{DVSGesture} &  \textbf{Cifar10-DVS}\\
\hline
SpikGRU    & 98.10\% & 66\% & 26.53\% \\
CS-GRU (ours)   & \textbf{99.31\%} & \textbf{82\%} & \textbf{51.40}\%\\
\hline
\end{tabular}
\end{table}

The results in the table show that CS-GRU performs better than SpikGRU in all the datasets tested. In MNIST, \mbox{CS-GRU} achieves a marginally higher accuracy. The performance gap increases with DVSGesture and CIFAR10-DVS, both of which are event-based and exhibit complex spatiotemporal features. For DVSGesture, CS-GRU exhibits significantly higher recognition rates, attributing to its enhanced ability to perceive dynamic nuances of gesture data. Meanwhile, on the more challenging CIFAR10-DVS dataset, CS-GRU far surpasses SpikGRU, reflecting its better capacity for dealing with and processing rapidly changing, high-dimensional inputs. These findings underscore the value of the architectural innovations in CS-GRU, namely its advanced gating mechanisms, which enable improved temporal integration of features in spiking neural networks.

\subsection{Energy efficiency}
To quantify the energy‐saving benefits of convolutional operations in inclusion with the Spiking GRU architecture, we compare the spiking (firing) activity rates of our CS-GRU and of SpikGRU for the same workload (DVS Gesture dataset).
The spiking activity rate is here defined as the average percentage of neurons that do emit a spike at every time step—a crude measurement of how "busy" your network is. As modern neuromorphic hardware only consumes substantial power when it is processing spikes (that is, inactive neurons consume no power), a lower spiking rate translates onto lower energy consumption directly. That is, by generating fewer, more meaningful spikes, a network not only performs less redundant computation, but also preserves battery life and decreases heat generation -- making spiking activity rate an easy and effective proxy for energy efficiency in general. Our CS-GRU produces only 
$0.13$ spikes per neuron per time step, about a $69.05$\% reduction in spiking activity as compared to the SpikGRU’s $0.42$ spikes per neuron per time step.
\begin{table}[h]
\centering
\begin{tabular}{l c c}
\hline
\textbf{Model}  & Spiking activity rate & Relative Reduction \\
                & [spikes/neuron/timestep] & [\%] \\
\hline
SpikGRU & $0.42$ & --                  \\
\hline
CS-GRU  & $0.13$ & \(69.05\%\downarrow\) \\
\hline
\end{tabular}
\caption{Comparison of average spiking activity rates for SpikGRU and CS-GRU.}
\end{table}

\section{CONCLUSION AND FUTURE WORKS}
\label{sec:conclusion}
We introduced a new building block in this paper, the Convolutional Spiking GRU (CS-GRU), that combines the strengths of convolutional computation, spiking neural dynamics, and GRU gating mechanisms. The large-scale experimentation on both sequential and spatio-temporal data sets demonstrates that the CS-GRU learns effective abstract representations as well as enhanced performance over a broad spectrum of data types. The empirical results confirm that CS-GRU leverages localized feature extraction and long-term temporal dynamics, overcoming the limitations in traditional recurrent and spiking neural models. The promising performance of \mbox{CS-GRU} suggests that it can be a good candidate building block for advanced spiking network models, including ensemble models, ResNets, or VGG networks. Open research directions encompass the  integration of the CS-GRU into higher-scale networks and its use on more diverse tasks, including speech recognition, video processing, and sensor data processing.

\bibliographystyle{IEEEbib}
\bibliography{strings,refs}

\begin{thebibliography}{10}

\bibitem{dampfhoffer2022investigating}
Manon Dampfhoffer, Thomas Mesquida, Alexandre Valentian, and Lorena Anghel,
\newblock ``Investigating current-based and gating approaches for accurate and
  energy-efficient spiking recurrent neural networks,''
\newblock in {\em International Conference on Artificial Neural Networks}.
  Springer, 2022, pp. 359--370.

\bibitem{maass1997networks}
Wolfgang Maass,
\newblock ``Networks of spiking neurons: the third generation of neural network
  models,''
\newblock {\em Neural networks}, vol. 10, no. 9, pp. 1659--1671, 1997.

\bibitem{gerstner2002spiking}
Wulfram Gerstner and Werner~M Kistler,
\newblock {\em Spiking neuron models: Single neurons, populations, plasticity},
\newblock Cambridge university press, 2002.

\bibitem{hausser2000hodgkin}
Michael H{\"a}usser,
\newblock ``The hodgkin-huxley theory of the action potential,''
\newblock {\em Nature neuroscience}, vol. 3, no. 11, pp. 1165--1165, 2000.

\bibitem{brunel2007lapicque}
Nicolas Brunel and Mark~CW Van~Rossum,
\newblock ``Lapicque’s 1907 paper: from frogs to integrate-and-fire,''
\newblock {\em Biological cybernetics}, vol. 97, no. 5, pp. 337--339, 2007.

\bibitem{brette2005adaptive}
Romain Brette and Wulfram Gerstner,
\newblock ``Adaptive exponential integrate-and-fire model as an effective
  description of neuronal activity,''
\newblock {\em Journal of neurophysiology}, vol. 94, no. 5, pp. 3637--3642,
  2005.

\bibitem{de2024spike}
Florent De~Geeter, Damien Ernst, and Guillaume Drion,
\newblock ``Spike-based computation using classical recurrent neural
  networks,''
\newblock {\em Neuromorphic Computing and Engineering}, vol. 4, no. 2, pp.
  024007, 2024.

\bibitem{gerstner2014neuronal}
Wulfram Gerstner, Werner~M Kistler, Richard Naud, and Liam Paninski,
\newblock {\em Neuronal dynamics: From single neurons to networks and models of
  cognition},
\newblock Cambridge University Press, 2014.

\bibitem{medsker2001recurrent}
Larry~R Medsker, Lakhmi Jain, et~al.,
\newblock ``Recurrent neural networks,''
\newblock {\em Design and Applications}, vol. 5, no. 64-67, pp. 2, 2001.

\bibitem{dey2017gate}
Rahul Dey and Fathi~M Salem,
\newblock ``Gate-variants of gated recurrent unit (gru) neural networks,''
\newblock in {\em 2017 IEEE 60th international midwest symposium on circuits
  and systems (MWSCAS)}. IEEE, 2017, pp. 1597--1600.

\bibitem{neftci2019surrogate}
Emre~O Neftci, Hesham Mostafa, and Friedemann Zenke,
\newblock ``Surrogate gradient learning in spiking neural networks: Bringing
  the power of gradient-based optimization to spiking neural networks,''
\newblock {\em IEEE Signal Processing Magazine}, vol. 36, no. 6, pp. 51--63,
  2019.

\bibitem{anumula2018feature}
Jithendar Anumula, Daniel Neil, Tobi Delbruck, and Shih-Chii Liu,
\newblock ``Feature representations for neuromorphic audio spike streams,''
\newblock {\em Frontiers in neuroscience}, vol. 12, pp. 23, 2018.

\bibitem{cramer2020heidelberg}
Benjamin Cramer, Yannik Stradmann, Johannes Schemmel, and Friedemann Zenke,
\newblock ``The heidelberg spiking data sets for the systematic evaluation of
  spiking neural networks,''
\newblock {\em IEEE Transactions on Neural Networks and Learning Systems}, vol.
  33, no. 7, pp. 2744--2757, 2020.

\bibitem{kingma2015adam}
P~Kingma~Diederik and Ba~Jimmy,
\newblock ``Adam: A method for stochastic optimization. 2014,''
\newblock {\em URL https://arxiv. org/abs/1412.6980}, 2015.

\bibitem{kim2022rate}
Youngeun Kim, Hyoungseob Park, Abhishek Moitra, Abhiroop Bhattacharjee,
  Yeshwanth Venkatesha, and Priyadarshini Panda,
\newblock ``Rate coding or direct coding: Which one is better for accurate,
  robust, and energy-efficient spiking neural networks?,''
\newblock in {\em ICASSP 2022-2022 IEEE International Conference on Acoustics,
  Speech and Signal Processing (ICASSP)}. IEEE, 2022, pp. 71--75.

\bibitem{amir2017low}
Arnon Amir, Brian Taba, David Berg, Timothy Melano, Jeffrey McKinstry, Carmelo
  Di~Nolfo, Tapan Nayak, Alexander Andreopoulos, Guillaume Garreau, Marcela
  Mendoza, et~al.,
\newblock ``A low power, fully event-based gesture recognition system,''
\newblock in {\em Proceedings of the IEEE conference on computer vision and
  pattern recognition}, 2017, pp. 7243--7252.

\bibitem{li2017cifar10}
Hongmin Li, Hanchao Liu, Xiangyang Ji, Guoqi Li, and Luping Shi,
\newblock ``Cifar10-dvs: an event-stream dataset for object classification,''
\newblock {\em Frontiers in neuroscience}, vol. 11, pp. 309, 2017.

\end{thebibliography}

\end{document}